\documentclass[10pt,twocolumn,letterpaper]{article}

\usepackage{3dv}
\usepackage{times}
\usepackage{epsfig}
\usepackage{graphicx}
\usepackage{amsmath}
\usepackage{amssymb}
\usepackage{caption}

\usepackage[pagebackref=true,breaklinks=true,colorlinks,bookmarks=false]{hyperref}

\threedvfinalcopy 

\def\threedvPaperID{436} 

\ifthreedvfinal\fi
\begin{document}

\title{Adaptive Learning for Multi-view Stereo Reconstruction}


\author{Qinglu Min$^{1}$, Jie Zhao$^{2}$, Zhihao Zhang$^{3}$, Chen Min$^{4}$\footnotemark[1]\\
	$^{1}$Jinggangshan University  $^{2}$Jinan Vocational College $^{3}$Tongji University $^{4}$Peking University\\
	{\tt\small minchen@stu.pku.edu.cn}
}
\maketitle

\footnotetext[1]{Corresponding author.}
\begin{abstract}
   Deep learning has recently demonstrated its excellent performance on the task of multi-view stereo (MVS). However, loss functions applied for deep MVS are rarely studied. In this paper, we first analyze existing loss functions' properties for deep depth based MVS approaches. Regression based loss leads to inaccurate continuous results by computing mathematical expectation, while classification based loss outputs discretized depth values. To this end, we then propose a novel loss function, named adaptive Wasserstein loss, which is able to narrow down the difference between the true and predicted probability distributions of depth. Besides, a simple but effective offset module is introduced to better achieve sub-pixel prediction accuracy. Extensive experiments on different benchmarks, including DTU, Tanks and Temples and BlendedMVS, show that the proposed method with the adaptive Wasserstein loss and the offset module achieves state-of-the-art performance.
\end{abstract}

\section{Introduction}

\begin{figure*}[h]
	\centering
	\includegraphics[width=6 in]{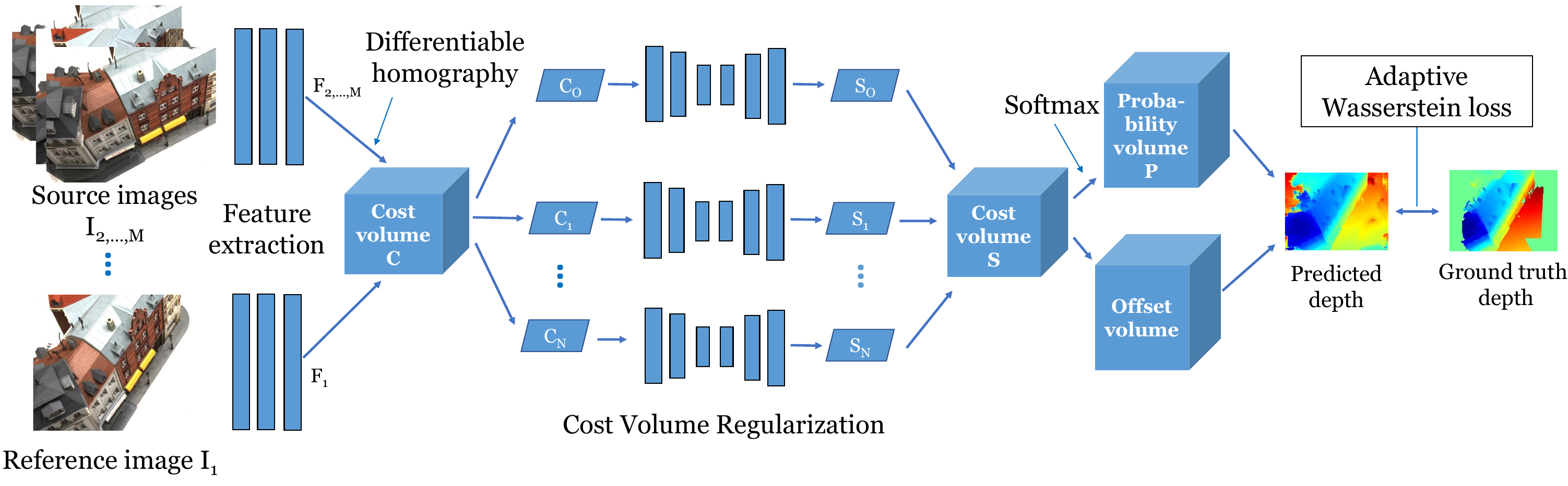}
	\caption{An overview of our proposed method. It simultaneously predicts a probability for the fixed discrete depth value and an additional offset value for each discrete depth value. The continuous depth values can be obtained by adding them together. This simple but effective offset module helps to improve the discrete depth prediction to achieve sub-pixel depth accuracy. The model is trained end-to-end by the adaptive Wasserstein loss for the predicted depth and ground depth distributions may not have any common supports.}
	\label{fig:flowchart}
\end{figure*} 

Multi-view stereo (MVS) seeks to recover 3D dense representation from a set of multi-view images with given calibrated cameras, which is a fundamental problem in computer vision. It is essential to a variety of applications such as augmented reality, 3D modeling and autonomous driving, and has been extensively studied for decades. While traditional MVS methods have achieved impressive results with hand-crafted photo-consistency metrics (e.g. SSD, NCC), they are still incompetent on low-texture regions, illumination changes and reflections which make dense matching intractable for accurate and complete 3D reconstruction. Recent learning based approaches~\cite{mvsnet, rmvsnet, pmvsnet, fastmvsnet, pvamvsnet} with deep CNNs have achieved considerably better performance and become one of the mainstream approaches for MVS. For instance, MVSNet~\cite{mvsnet} encodes camera geometries in the network to build 3D cost volumes upon CNN features and uses 3D CNNs for cost volume regularization. MVSNet and its variants~\cite{mvscrf,d2hcrmvsnet,rmvsnet,ucsnet,casmvsnet} significantly improve the accuracy and completeness of MVS reconstruction compared to traditional methods based on hand-crafted metrics. 

A key part of learning based approaches is to define a loss function to obtain a better optimization objective and learning representation from training data~\cite{wingloss,adaptivewingloss}. However, the community of learning based MVS seems to pay little attention on loss function design. To our knowledge, adopted loss functions of most existing learning based MVS approaches can be categorized into two types: regression based~\cite{fastmvsnet,casmvsnet,cvpmvsnet,patchmatchnet,attmvs} and classification based~\cite{rmvsnet,d2hcrmvsnet}. The pioneering MVSNet first predicts a probability distribution of depth on a pre-defined set of discretized depth values and then calculates the continuous \textit{expectation} of depth from this distribution. The final loss is computed by L1 loss. Though achieving promising results, MVSNet and its followers~\cite{mvsnet,mvscrf,casmvsnet,pointmvsnet,cvpmvsnet,ucsnet} may predict wrong depth as the predicted depth distribution is likely to have more than one peaks but the depth \textit{expectation} will be far from any predicted peaks (more details in Fig.~\ref{fig:loss}(c)). While R-MVSNet~\cite{rmvsnet} and D$^{2}$HC-RMVSNet~\cite{d2hcrmvsnet}, designed for wide depth range scenarios with recurrent neural network, treat the depth learning task as a classification problem with pixel-wise cross entropy loss where each depth hypothesis is considered as a class. However, the outputs of classification based methods are discretized so that sub-pixel accuracy is hard to obtain as shown in Fig.~\ref{fig:loss}(d).

To address the aforementioned issues, inspired by the idea of Wasserstein distance and offset learning used in CDN~\cite{wassersteinstereo} for stereo disparity estimation, we build a novel neural network architecture for deep depth based MVS that is able to output the continuous depth values as shown in Fig.~\ref{fig:flowchart}. It both predicts a probability for the fixed discrete depth values like the classification based MVS methods and an offset value for each discrete depth value. It takes the \textit{mode} from the probability of the fixed discrete depth values and adds an offset value to the \textit{mode} to obtain a continuous depth value, as shown in Fig.~\ref{fig:loss}(e). With the simple offset module, the output is transformed from a discrete distribution to a continuous distribution. 

Next, we introduce the adaptive Wasserstein loss for deep depth based MVS. The learning goal of deep depth based MVS is to minimize the divergence between the predicted depth distribution and the ground truth depth distribution (i.e. the Dirac delta distribution). One concern is that the predicted depth and the ground truth depth distributions may not have any common supports, where the Kullback-Leibler divergence used in the existing classification based MVS methods will be invalid. Therefore, we introduce the Wasserstein metric~\cite{wassersteindistance} to deep depth based MVS, which is efficient in measuring the divergence between two distributions that do not have common supports.

We conduct extensive experiments on DTU~\cite{dtu}, Tanks and Temples~\cite{tanks} and BlendedMVS~\cite{blendedmvs} datasets for performance evaluation. The proposed method achieves the state-of-the-art performance over many existing methods on MVS. Besides, we validate our approach by thorough studies to demonstrate the benefits of the introduced offset module and the adaptive Wasserstein loss, offering more insight into its effectiveness for deep depth based MVS.

In summary, our \textbf{main contributions} include:

\begin{itemize}
	\item We propose a simple but effective offset module for learning based MVS that is capable of yielding continuous depth values.  
	\item We introduce the adaptive Wasserstein loss for learning based MVS, which is able to narrow down the divergence between the true and the predicted depth distributions that may not have any common supports.
	\item Our approach achieves state-of-the-art performance
	over many existing methods for MVS on the DTU and Tanks and Temples benchmark datasets.
\end{itemize}

\section{Related Work}

\subsection{Learning based MVS}

Recently, learning based methods have shown great potentials for MVS reconstruction. SurfaceNet~\cite{surfacenet} and LSM~\cite{lsm} first construct a cost volume with multi-view images and then use the 3D CNNs to regularize and infer the voxel of the volume. However, they can only reconstruct small scale scenes for the limitation of volumetric based methods. In contrast, MVSNet~\cite{mvsnet} focuses on producing the depth map for one reference image at each time, which can adaptively reconstruct a large scene directly. 
In the last three years, many variants of MVSNet have been proposed and they can be divided into one stage methods~\cite{mvscrf,cider,rmvsnet,d2hcrmvsnet,fastmvsnet} and coarse-to-fine methods~\cite{pointmvsnet,casmvsnet,ucsnet,cvpmvsnet,attmvs,vismvsnet} according to the way of learning depth. One stage methods, such as MVSCRF~\cite{mvscrf} with the conditional random fields (CRFs), P-MVSNet~\cite{pmvsnet}  with the pixel-wise correspondence information aggregation, have been proposed to improve MVSNet.
R-MVSNet~\cite{rmvsnet} and $D^2$HC-RMVSNet~\cite{d2hcrmvsnet} sequentially regularize the 2D cost maps along the depth direction via GRU or LSTM to reduce the memory consumption. In order to refine the depth map, some coarse-to-fine methods have been proposed. Point-MVSNet~\cite{pointmvsnet} proposes to use a small cost volume to generate the coarse depth and uses a point-based iterative refinement network to output the full resolution depth maps. CasMVSNet~\cite{casmvsnet}, UCSNet~\cite{ucsnet}, CVP-MVSNet~\cite{cvpmvsnet}, AttMVS~\cite{attmvs}, Vis-MVSNet~\cite{vismvsnet} and PatchmatchNet~\cite{patchmatchnet} first process a small standard plane sweep volume to predict low-resolution depth and then narrow the depth range of each stage by the prediction from the previous stage. Although the coarse-to-fine methods can predict accurate depths, they adopt the L1 or smooth L1 loss to predict the \textit{expectation} values of depth prediction which may lead to incorrect results for multi peak predictions. We propose a different coarse-to-fine method that classifies the depth values into sampled fixed depths and refines the depths with the additional offset values. 

\subsection{Offset Learning}

The design of sub-network for offset prediction has been proven to be effective in many computer vision tasks~\cite{ssd,yolo,wassersteinstereo,centernet}. G-RMI~\cite{g-rmi} predicts both the heatmaps at fixed locations and the offsets for each keypoint for multi-person pose estimation in the wild. CDN~\cite{wassersteinstereo} starts with the integral disparity values and predicts offsets in addition to disparity probabilities for accurate disparity estimation. 
The anchor-free object detection methods CenterNet~\cite{centernet} and CenterPoint~\cite{centerpoint} use keypoint estimation to find center points and additionally predicts a local offset for each center point in order to recover the discretization error caused by the output stride. In this paper, we introduce the offset module to the deep depth based MVS for continuous depth estimation. 

\subsection{Wasserstein Distance}

Wasserstein distance or Earth-Mover (EM) distance~\cite{wassersteindistance} measures the minimal effort required to reconfigure the probability mass of one distribution in order to recover the other distribution inspired by the problem of optimal mass transportation. In particular, Wasserstein Auto-Encoder~\cite{wassersteinautoencoder} minimizes a penalized form of the Wasserstein distance between the model distribution and the target distribution, and Wasserstein GAN~\cite{wassersteingan} applies the Wasserstein-$1$ metric to alleviate the vanishing gradient and the mode collapse issues of Generative Adversarial Networks (GAN)~\cite{gan}. In this paper, we introduce the Wasserstein distance to learning based MVS to narrow the divergence between the true and predicted depth distributions that may not have any common supports. 

\section{Method}

\subsection{Review of learning based MVS}
Learning based MVS methods usually consist of five stages: pre-processing, feature extraction, cost volume building, cost volume regularization and post-processing~\cite{mvscrf}. 

The images are first processed by traditional Structure from Motion (SfM) methods like Colmap~\cite{colmap} or OpenMVG~\cite{openmvg} to get the camera intrinsics and extrinsics, while the depth ranges and view selections of multiple images are obtained from sparse reconstruction. And then the reference image $\mathbf{I}_1\in \mathbb{R}^{H\times W\times 3}$ and a set of its neighboring source images $\left \{{\mathbf{I}_i}\right \}_{i=2}^{M}\in \mathbb{R}^{H\times W\times 3}$ are fed into the 2D feature extraction network to extract the 2D deep features $\left \{{\mathbf{F}_i}\right \}_{i=1}^{M}\in \mathbb{R}^{H\times W\times C}$. Next, these 2D image features are warped into the reference camera frustum by differentiable homographies to build the 3D feature volumes $\left \{{\mathbf{V}_i}\right \}_{i=2}^{M}$. All the 3D feature volumes are merged by a variance based cost metric to one 4D cost volume $\mathbf{C}\in \mathbb{R}^{H\times W \times N \times C}$. 3D CNNs in~\cite{mvsnet,casmvsnet,cvpmvsnet,attmvs} and recurrent neural network in~\cite{rmvsnet,d2hcrmvsnet} are used to regularize the 4D cost volume $\mathbf{C}$ and output a 3D cost volume $\mathbf{S}\in \mathbb{R}^{H\times W \times N}$. \textit{Softmax} operation is applied to convert the cost volume $\mathbf{S}$ to a 3D probability volume $\mathbf{P}\in \mathbb{R}^{H\times W \times N}$. In order to produce the depth map $\mathbf{D}\in \mathbb{R}^{H\times W}$, there are two mainstream  options: (1) Regression based methods: one can perform \textit{soft argmin} operation on $\mathbf{P}$ and compute the \textit{expectation} value along depth direction as the depth value~\cite{gcnet}. (2) Classification based methods: the depth map can be obtained by just taking \textit{argmax} operation of winner-take-all on $\mathbf{P}$ to select the \textit{mode} as the depth value~\cite{rmvsnet}. Finally, point clouds are obtained by filtering the depth map with the photo-metric and the geometric consistencies. 

\subsection{Analysis of Existing Loss Functions}

The goal of most learning based MVS is to determine the depth values of the reference image pixels. For one image pixel, its depth value is within the given depth range $[d_{min},d_{max}]$. According to the way to obtain the depth, the existing losses for learning based MVS can be divided into regression based and classification based. 

\subsubsection{Regression based Loss}

The pioneering deep depth based MVS method MVSNet~\cite{mvsnet} and some of its variants~\cite{mvscrf,cider,casmvsnet,patchmatchnet} regard the depth learning problem as a regression task. In order to obtain the continuous depth values, they adopt the \textit{soft argmin} operation~\cite{gcnet} on the probability volume $\mathbf{P}$ to get the \textit{expectation} depth map $\mathbf{D}$ along the depth direction~\cite{mvsnet}:
\begin{equation} \label{soft_argmin}
\mathbf{D} = \sum_{d_s=d_{min}}^{d_{max}}d_s \cdot \mathbf{P}(d_s),
\end{equation}
where $d_s$ is the sampled fixed discrete depth value in depth range $[d_{min}, d_{max}]$ and the number of sampled $d_s$ is $N$, $\mathbf{P}\in \mathbb{R}^{H\times W \times N}$ is the 3D probability volume, and $\mathbf{D}\in \mathbb{R}^{H\times W}$ is the predicted depth map as stated above. While the depth hypotheses are uniformly sampled within the depth range $[d_{min}, d_{max}]$ during cost volume construction, the \textit{expectation} value here is able to produce a continuous depth estimation.

The MVS networks with regression based loss can be trained end-to-end using the L1 or smooth L1 loss to regress the depths~\cite{mvsnet}:
\begin{equation} \label{l1_loss}
loss_{reg} = \sum_{(u,v)\in \mathbf{P}_{valid}}\left \| \mathbf{D}(u,v)-\hat{\mathbf{D}}(u,v)\right \|_1,
\end{equation}
where $\hat{\mathbf{D}}\in \mathbb{R}^{H\times W}$ is the ground truth depth map, and $\mathbf{P}_{valid}$ denotes the set of valid ground truth pixels.

Though the regression based loss with the \textit{soft argmin} operation yields continuous depth values, the \textit{expectation} values may be wrong when the predicted probability is multi-modal distribution for the challenging regions such as low-textured planes, occluded areas and boundary areas where the pixels are easy to be mismatched. As shown in Fig.~\ref{fig:loss}(c), the \textit{expectation} value in multi-modal distribution is not aligned with any peaks. The other problem with regression based loss is that it is only valid if depth values are uniformly sampled within the depth ranges, while it is not suitable for \textit{inverse depth} sampling strategy (i.e. the discrete depth hypotheses are uniformly sampled in inverse depth space in order to  make the 2D points that lie in the same epipolar line distribute as uniformly as possible~\cite{rmvsnet}.) that is designed for wide depth range reconstructions~\cite{cider}. To this end, the classification based loss for MVS has been proposed. 

\subsubsection{Classification based Loss}

The pre-assumption of regression based loss is that the depth hypotheses are uniformly sampled within depth range to calculate the \textit{expectation} depth values. However, it is not suitable for wide depth range reconstructions. Thus some deep depth based MVS methods do not treat the MVS problem as a regression task, they train the network as a classification problem with \textit{inverse depth} setting~\cite{rmvsnet}. They use the cross entropy loss to train the network end-to-end:
\begin{equation} \label{Cross_Entropy_loss}
loss_{cla} = \sum_{(u,v)\in \mathbf{P}_{valid}}\sum_{d_s=d_{min}}^{d_{max}}-\mathbf{P}(u,v,d_s)\cdot log \mathbf{Q}(u,v,d_s),
\end{equation}
where $\mathbf{P}(u,v,d_s)$ is a voxel in $\mathbf{P}$, $\mathbf{Q}\in \mathbb{R}^{H\times W\times N}$ is the ground truth binary occupancy volume which is generated by the one hot encoding of the ground truth depth map $\hat{\mathbf{D}}$ and $\mathbf{Q}(u,v,d_s)$ is the corresponding voxel to $\mathbf{P}(u,v,d_s)$.

With the \textit{inverse depth} sampling strategy, the MVS methods with classification based loss can efficiently handle reconstruction with wide depth range. However, its disadvantage is also obvious: the output of the network is a set of fixed discrete depth values, thus it can not achieve sub-pixel accuracy. As shown in Fig.~\ref{fig:loss}(d), the predicted discrete depth value with the classification based loss deviates from the ground truth depth. To alleviate this problem, R-MVSNet~\cite{rmvsnet} proposes a refinement algorithm to refine the depth map in a small depth range by enforcing the multi-view photo-consistency. But this refinement algorithm is time-consuming (i.e. R-MVSNet takes 2.9s to infer the initial depth map and 6.2s to perform the depth map refinement~\cite{rmvsnet}.). 

\subsection{Offset Learning} 

We propose a new neural network architecture for deep depth based MVS, which can not only handle wide depth ranges but also output continuous depth values. 
We add an additional offset module for deep depth based MVS:
\begin{equation} \label{offset}
d^{'} = d_s +\mathbf{Offset}(u,v,d_s),
\end{equation}
where $\mathbf{Offset}\in \mathbb{R}^{H\times W\times N}$ is the 3D offset volume. $\mathbf{Offset}(u,v,d_s)$ is a voxel in $\mathbf{Offset}$ and corresponding to the fixed depth probability $\mathbf{P}(u,v,d_s)$. 

The model with offset module first takes the \textit{mode} from the probability of the fixed discrete depth values and then adds an offset value to the \textit{mode} to obtain a continuous depth value. With the offset module, the output is transformed from a discrete distribution to a continuous distribution. The sub-network for offset prediction is two 3D CNN blocks added to the end of regularization part of deep depth based MVS methods and shares the feature learning and cost volume regularization with the fixed depth probability learning as shown in Fig.~\ref{fig:flowchart}. 

We can simply add the offset module to the classification based MVS methods~\cite{rmvsnet,d2hcrmvsnet} to obtain the continuous depth values. However, the cross entropy loss (i.e. the Kullback-Leibler divergence) used in the classification based MVS methods will be invalid when the predicted depth distribution and the ground truth depth distribution do not have any commom supports. Thus, we propose the Wasserstein Loss for deep depth based MVS. 

\subsection{Adaptive Wasserstein Loss} 

The distribution of ground truth depth $\mathbb{D}$ is the Dirac delta distribution. The goal of deep depth based MVS methods is to learn to minimize the divergence between the ground truth distribution $\mathbb{D}$ and predicted depth distribution $\mathbb{P}$. 
One concern is that the predicted depth distribution and the ground truth depth distribution may not have any common supports, thus we apply the Wasserstein distance~\cite{wassersteindistance} to measure the predicted depth distribution and the ground truth depth distribution. The Wasserstein-$p$ distance or Earth-Mover (EM) distance is defined as:
\begin{equation} \label{Wasserstein_distance}
W(\mathbb{P},\mathbb{D})=\inf_{\gamma \in \Pi (\mathbb{P},\mathbb{D})}\mathbb{E}_{(x,y)\sim \gamma}\begin{bmatrix}
\left \| x-y \right \|^{p}
\end{bmatrix}^{1/p},
\end{equation} 
where $\Pi (\mathbb{P},\mathbb{D})$ denotes the set of all joint distributions $\gamma(x,y)$ whose marginals are respectively $\mathbb{P}$ and $\mathbb{D}$. Intuitively, $\gamma(x,y)$ indicates how much “mass” must be transported from $x$ to $y$ in order to transform the distribution $\mathbb{P}$ into the distribution $\mathbb{D}$. The EM distance then is the “cost” of the optimal transport plan~\cite{wassersteingan}.

For the task of deep depth based MVS, the Wasserstein distance measures the minimal effort required to reconfigure the probability mass of probability distribution of predicted depth $\mathbb{P}$ in order to recover the probability distribution of ground truth depth $\mathbb{D}$. Calculating the Wasserstein distance for the probability distributions of multi-dimensional variables is very tricky as a linear programming problem. Fortunately, the depth values of learning based MVS are the one-dimensional variables~\cite{wassersteinstereo}, thus we can simplify the Wasserstein-$p$ distance for deep depth based MVS as: 
\begin{equation} \label{Wasserstein_distance2}
	\begin{aligned}
	W(\mathbb{P},\mathbb{D})&=\inf_{\gamma \in \Pi (\mathbb{P},\mathbb{D})}\mathbb{E}_{(d^{'},\hat{d})\sim \gamma}\begin{bmatrix}
	\left \| d^{'}-\hat{d} \right \|^{p}
	\end{bmatrix}^{1/p} \\&=\left (\mathbb{E}_{\mathbb{P}}\mathbb{E}_{\mathbb{D}}\left \| d^{'}-\hat{d} \right \|^p\right )^{1/p}, 
	\end{aligned}
\end{equation} 		
where $d^{'}$ is the predicted depth value and $\hat{d}$ is the corresponding ground truth depth value. As the probability distribution of ground truth depth $\mathbb{D}$ is a Dirac delta distribution, the Wasserstein-$p$ distance for deep depth based MVS can be rewritten as:
\begin{equation} \label{Wasserstein_distance3}
\begin{aligned}
W(\mathbb{P},\mathbb{D})=\left (\mathbb{E}_{\mathbb{P}}\left \| d^{'}-\hat{d} \right \|^p\right )^{1/p}. 
\end{aligned}
\end{equation} 		
With the predicted continuous depth value $d^{'}$ by the offset module, we can rewrite the Wasserstein-$p$ distance $W(\mathbb{P},\mathbb{D})$ as:
\begin{equation} \label{Wasserstein_distance4}
	\begin{aligned}
	W_{dis}(u,v) =\left (\sum_{d_s=d_{min}}^{d_{max}}\mathbf{P}(u,v,d_s)\left  \|\Delta d\right \|^p\right)^{1/p},
	\end{aligned}
\end{equation} 		
\begin{equation} \label{Wasserstein_distance4}
\begin{aligned}
&\Delta d = d_s+  \mathbf{Offset}(u,v,d_s)-\mathbf{Q}(u,v,d_s).
\end{aligned}
\end{equation} 		
Calculating the Wasserstein-$p$ distance for all the valid pixels, we can define Wasserstein loss for deep depth based MVS as:
\begin{equation} \label{Wasserstein_Loss5}
loss_{was} = \sum_{(u,v)\in \mathbf{P}_{valid}} W_{dis}(u,v).
\end{equation}

In order to make sure that the predicted depths peak at the ground truth depths, we also add the classification loss to the model to adaptively focus on the \textit{mode}. The adaptive Wasserstein loss fuction is defined as:
\begin{equation} \label{loss_all}
loss_{ada\_was} = loss_{was} + loss_{cla}.
\end{equation}

Our adaptive Wasserstein loss with offset module is both mathematically well founded and effective. Compared with the regression based loss that computes the \textit{expectation} value along depth direction as the depth value, the proposed adaptive Wasserstein loss with the offset module uses the \textit{mode} as the prediction instead of the \textit{expectation} during inference, so as to ensure that the predicted depth has a high estimation probability. Compared with the classification based loss that outputs fixed discreted depth values, our Wasserstein loss with offset module can well measure the divergence between the ground truth distribution and predicted depth distribution and output continuous depth values. 

\subsection{Network Architecture}

We follow the basic network architecture of $D^2$HC-RMVSNet~\cite{d2hcrmvsnet} for the state-of-the-art performance of wide depth range prediction, which uses the lightweight feature extractor DRENet and hybrid recurrent regularization module HU-LSTM. We add the offset module to the end of the regularization module and replace the original cross entropy loss with our adaptive Wasserstein loss. 

\section{Experiments and Analyses}	

In this section, a series of experiments are conducted to validate the performance of the proposed method.

\subsection{Datasets}

\subsubsection{DTU Dataset}
DTU dataset~\cite{dtu} is a most widely used dataset for learning based MVS, which composes of 124 different scenes and each scene has 49 or 64 images of resolution 1600$\times$1200 pixels. DTU dataset is an indoor dataset and acquired by a structured light scanner. It is split into training set (79 scenes), validation set (18 scenes) and evaluation set (22 scenes). Scenes in DTU dataset have different material, texture and geometric property. The ground truth depth maps are generated from the point clouds following MVSNet~\cite{mvsnet} but are not always complete. 

\subsubsection{Tanks and Temples Dataset}
The Tanks and Temples dataset \cite{tanks} is a large-scale outdoor benchmark collected in real-world environments. Its intermediate set consists of: Family, Francis, Horse, Lighthouse, M60, Panther, Playground, and Train. The Tanks and Temples dataset has varying scales, surface reflection and exposure conditions. It provides an online leaderboard for MVS methods. Following MVSNet~\cite{mvsnet}, the camera parameters and depth range of Tanks and Temples dataset are computed by Colmap~\cite{colmap}. 

\subsubsection{BlendedMVS Dataset}
BlendedMVS datset~\cite{blendedmvs} is a recently published large-scale dataset providing sufficient training ground truth for learning based MVS. BlendedMVS dataset is created by applying a 3D reconstruction pipeline to recover high-quality textured meshes from images of well-selected scenes. Then these mesh models are rendered to color images and depth maps. It contains over 17k high-resolution images covering a variety of scenes and is split into 106 training scenes and 7 validation scenes. 

\subsection{Implementation Details}

\subsubsection{Training}

We train our method on DTU training set. The training image size is set to $W\times H = 160 \times 128$ and the number of reference views $N_v$ is 7. The depth hypotheses are sampled from $425mm$ to $935mm$ with depth plane number $N=192$. Source images for the given reference are selected as MVSNet. We implement our model with PyTorch~\cite{pytorch} and use the Adam~\cite{adam} optimizer with initial learning rate of 0.001 and decrease the learning rate by 0.9 every epoch. The batch size is set to 2 on 2 NVIDIA GTX 2080Ti GPU devices.

\subsubsection{Testing}
\begin{figure*}[h]
	\centering
	\includegraphics[width=4.8 in]{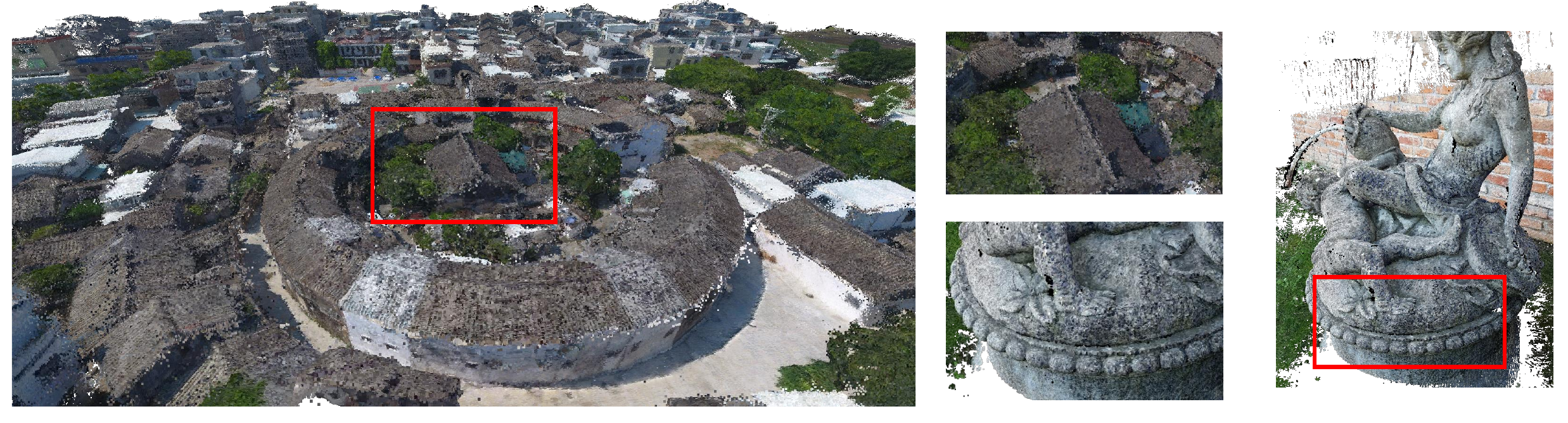}
	\caption{Reconstruction results of the validation set in the BlendedMVS dataset. Our method can reconstruct both small and large scale scenes, which demonstrates the reconstruction scalability of our method.}
	\label{fig:blendedmvs}
\end{figure*}
For testing, the number of reference views is set to $N_v
=7$ and depth planes number $N$ is set to $512$ for scalable MVS reconstruction. The input image resolution for evaluation is set to $800\times 600$ for DTU dataset, $1920\times 1056$ for Tanks and Temples dataset and $768\times 576$ for BlendedMVS dataset respectively. 

\subsubsection{Post-processing}

In order to fuse the learned depth maps for all the reference images into one single 3D dense point cloud, we first filter the depth maps with the photo-metric and the geometric consistency checking strategy~\cite{rmvsnet} and then fuse the filtered depth maps. The photo-metric constraint measures the matching quality. The probability score of the selected depth value is selected as its confidence measurement and the pixels with probability lower than $0.25$ are discarded. The geometric constraint measures the multi-view depth consistency. We project the reference image through the learned depth to other view and then reproject the image back to the reference. The dynamic geometric consistency checking method presented in~\cite{d2hcrmvsnet} is used to cross-filter original depth maps. For depth map fusion, the visibility-based depth map fusion~\cite{merrell2007real} and the mean average fusion~\cite{mvsnet} are used to improve the depth map quality and generate the final 3D point cloud.

\subsection{Results}

We first demonstrate the state-of-the-art performance of our method on the DTU and Tanks and Temples dataset, and then extend our method on BlendedMVS dataset to reconstruct the small and large scale scenes to investigate the practicality and scalability.
\subsubsection{Results on DTU Dataset}
\begin{table}
	\caption{Quantitative results of reconstruction quality on DTU evaluation set (lower is better). Our method achieves a comparable result with other state-of-the-art methods. Specially, our method outperforms $D^2$HC-RMVSNet~\cite{d2hcrmvsnet} in terms of both accuracy and completeness with the same network architecture.}	
	\begin{center}
		\setlength{\tabcolsep}{0.2mm}{
			\begin{tabular}{c| ccc}
				\hline
				Method &Acc.(mm) &Comp.(mm) &Overall(mm)\\ 
				\hline
				Colmap~\cite{colmap} &0.400 &0.664 &0.536\\ 
				MVSNet~\cite{mvsnet} &0.396 &0.527 &0.462\\ 
				R-MVSNet~\cite{rmvsnet}&0.385 &0.459 &0.422\\ 
				Point-MVSNet~\cite{pointmvsnet}&0.342 &0.411 &0.376\\ 
				CVP-MVSNet~\cite{cvpmvsnet} &{\bf 0.296} &0.406 &0.351\\ 
				UCSNet~\cite{ucsnet} & 0.338 &0.349 &{\bf0.344}\\ 
				CasMVSNet~\cite{casmvsnet} &0.325 &0.385 &0.355\\  
				AttMVS~\cite{attmvs} &0.383 &0.329 &0.356 \\
				Vis-MVSNet~\cite{vismvsnet}&0.369 & 0.361 & 0.365\\
				PatchmatchNet~\cite{patchmatchnet}&0.427 &{\bf 0.277} &0.352 \\
				$D^2$HC-RMVSNet~\cite{d2hcrmvsnet} &0.395 &0.378 &0.386\\ 
				\hline
				Ours &0.364 & 0.354 &0.359  \\
				\hline
		\end{tabular}}
		\label{tab:DTU}
	\end{center}
\end{table}
\begin{table*}
	\caption{Quantitative results on the Tanks and Temples benchmark. The evaluation metric is f-score (higher is better). Our method ranks $1^{st}$ among all published MVS methods, which demonstrates the generalizability of the proposed method.}	
	\begin{center}
		\setlength{\tabcolsep}{2.3mm}{
			\begin{tabular}{c|c|cccccccclccc}
				\hline
				Method &Mean &Family &Francis &Horse &Lighthouse &M60 &Panther &Playground &Train \\ 
				\hline
				Colmap~\cite{colmap}&42.14 &50.41 &22.25 &25.63 &56.43 &44.83 &46.97 &48.53 &42.04  \\ 
				MVSNet~\cite{mvsnet}&43.48 &55.99 &28.55 &25.07 &50.79 &53.96 &50.86 &47.90 &34.69  \\ 
				R-MVSNet~\cite{rmvsnet}&43.48 &55.99 &28.55 &25.07 &50.79 &53.96 &50.86 &47.90 &34.69  \\ 
				Point-MVSNet~\cite{pointmvsnet}&48.27 &61.79 &41.15 &34.20 &50.79 &51.97 &50.85 &52.38 &43.06  \\ 
				CVP-MVSNet~\cite{cvpmvsnet}&54.03 &76.50 &47.74 &36.34 &55.12 &57.28 &54.28 &57.43 &47.54  \\ 
				UCSNet~\cite{ucsnet}&54.83 &76.09 &53.16 &43.03 &54.00 &55.60 &51.49 &57.38 &47.89  \\ 
				CasMVSNet~\cite{casmvsnet}&56.84 &76.37 &58.45 &46.26 &55.81 &56.11 &54.06 &58.18 &49.51  \\
				ACMM~\cite{acmm}&57.27 &69.24 &51.45 &46.97 &63.20 &55.07 &57.64 &60.08 &54.48 \\
				AttMVS~\cite{attmvs}&60.05 &73.90 &{\bf62.58} &44.08 &{\bf64.88} &56.08 &59.39 &{\bf63.42} &{\bf56.06}  \\
				Vis-MVSNet~\cite{vismvsnet} &60.03 &{\bf77.40} &60.23 &47.07 &63.44 &{\bf62.21} &57.28 &60.54 &52.07  \\
				PatchmatchNet~\cite{patchmatchnet} &53.15 &66.99 &52.64 &43.24 &54.87 &52.87 &49.54 &54.21 &50.81  \\
				$D^2$HC-RMVSNet~\cite{d2hcrmvsnet}&59.20 &74.69 &56.04 &49.42 &60.08 &59.81 &59.61 &60.04 &53.92  \\
				\hline 
				Ours& {\bf60.53} &77.18 &57.96 &{\bf 52.55} &60.03 & 60.73 &{\bf 61.23} &60.28 &54.26  \\
				\hline
		\end{tabular}}
		\label{tab:TT-result}
	\end{center}
\end{table*}

\begin{figure*}[h]
	\centering
	\includegraphics[width=5.6 in]{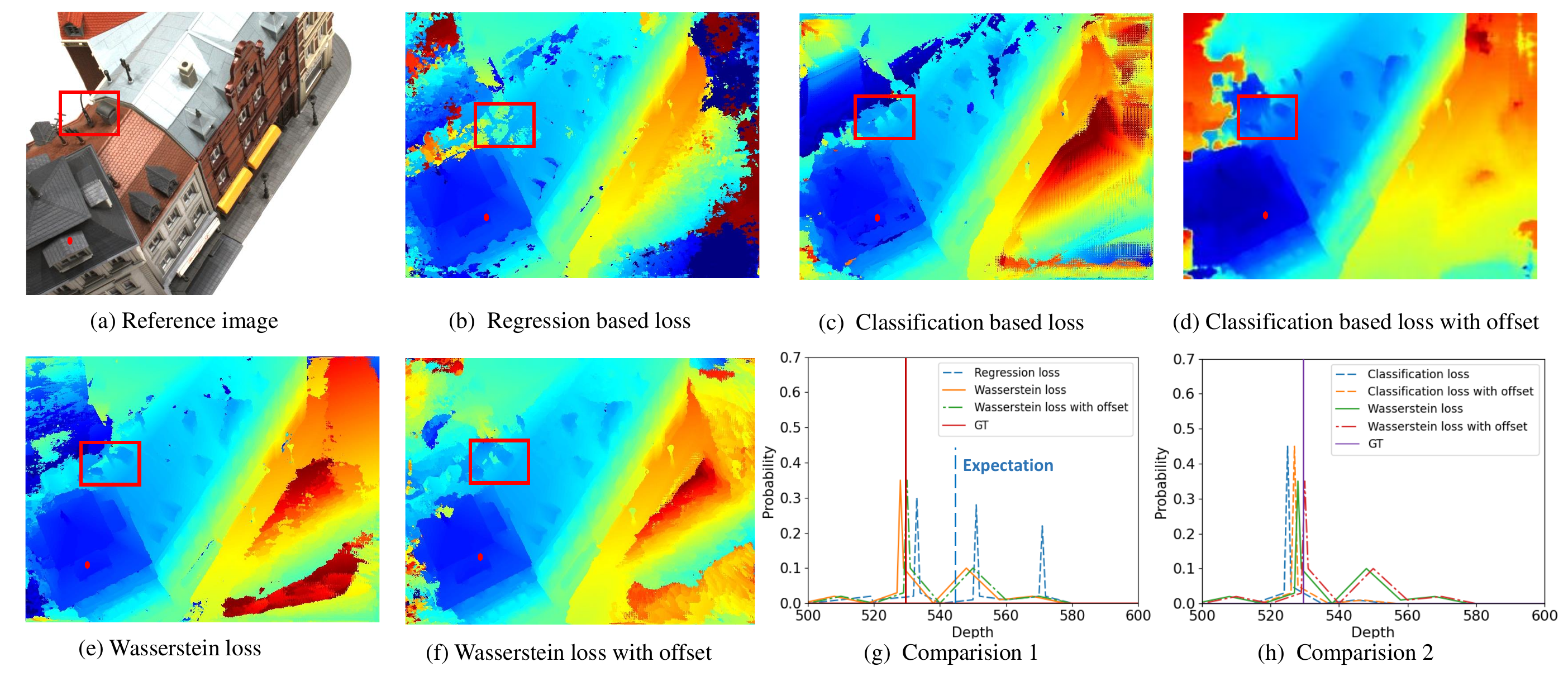}
	\caption{Comparision of different loss functions with and without the offset module on the inferred depth map of scan 9 in DTU dataset.}
	\label{fig:comparision}
\end{figure*}
We first evaluate our method on the DTU evaluation set using
the official matlab script provided by the DTU dataset~\cite{dtu}. We compute the overall score by the average of the accuracy and completeness to evaluate the overall reconstruction quality, where accuracy and completeness are the absolute distances between predicted and ground truth point clouds while overall is the mean average of the two metrics. The quantitative results are shown in Tab.~\ref{tab:DTU}. Our method achieves a overall score of 0.359, which is comparable with other state-of-the-art methods. 
It is noteworthy that our method and $D^2$HC-RMVSNet share the same network architecture, but our method produces more accurate and complete depth maps than $D^2$HC-RMVSNet. It proves the efficacy of our offset module and adaptive Wasserstein loss. 

\subsubsection{Results on Tanks and Temples Dataset}

Tanks and Temples dataset is a large-scale outdoor dataset captured in more complex environments, while the DTU dataset is an indoor dataset taken under well-controlled environment with fixed camera trajectory. To further demonstrate the generalization ability of our proposed method, we test our method on the Tanks and Temples dataset and the evaluation results are shown in Tab.~\ref{tab:TT-result}. 

Our method achieves a mean F-score of 60.53 and ranks $1^{st}$ among all published MVS methods on the intermediate set (date: July 31, 2021), which outperforms all classical MVS methods~\cite{colmap,acmm} and recent learning based approaches~\cite{mvsnet,rmvsnet,pointmvsnet,casmvsnet,cvpmvsnet,ucsnet,attmvs,patchmatchnet,d2hcrmvsnet,vismvsnet}. Please see the supplementary material for error visualization and reconstructed point clouds of our method.

\subsubsection{Results on BlendedMVS Dataset}

BlendedMVS dataset is a new large-scale MVS dataset covering a variety of scenes including cities, architectures, sculptures and small objects but does not officially provide the evaluation tools. Thus we test on it to further evaluate the practicality and scalability of our method, using trained model on the DTU dataset and fine-tuned on the BlendedMVS training set. As shown in Fig.~\ref{fig:blendedmvs}, our method can well reconstruct both small and large scale scenes.

\subsection{Ablation Study}
This section studies the effectiveness of our method for deep depth based MVS in detail. We need to investigate that: (1) Whether the adaptive Wasserstein loss can improve the performance of learning based MVS compared to the existing regression and classification based loss? (2) Whether the offset module can help to achieve sub-pixel accuracy?

\begin{table}
	\caption{Comparison of different loss functions on the DTU evaluation set. $L1$ refers to L1 loss, $CE$ refers to cross entropy loss and $W$ refers to the adaptive Wasserstein loss.}	
	\begin{center}
		\setlength{\tabcolsep}{3.2mm}{
			\begin{tabular}{c| ccc}
				\hline
				Loss &Acc.(mm) &Comp.(mm) &Overall(mm)\\ 
				\hline
				L1&0.423 &0.408 &0.416\\ 
				CE&0.385 &0.372 &0.379\\
				\hline
				W&{\bf 0.364} & {\bf 0.354} &{\bf 0.359}  \\
				\hline
		\end{tabular}}
		\label{tab:dif loss}
	\end{center}
\end{table}

\begin{table}
	\caption{Comparison of reconstruction results with and without the offset module on the DTU evaluation set.}	
	\begin{center}
		\setlength{\tabcolsep}{1.6mm}{
			\begin{tabular}{cc| ccc}
				\hline
				Loss&Offsets &Acc.(mm) &Comp.(mm) &Overall(mm)\\ 
				\hline
				W&$\times$&0.385 &0.368 &0.377\\ 
				W&$\checkmark$ &{\bf 0.364} & {\bf 0.354} &{\bf 0.359}  \\
				\hline	
				CE&$\times$&0.395 &0.378 &0.386\\ 
				CE&$\checkmark$ &0.385 &0.372 &0.379\\
				\hline
		\end{tabular}}
		\label{tab:offset}
	\end{center}
\end{table}

\subsubsection{Effectiveness of adaptive Wasserstein Loss}
Firstly, we compare the quantitative results of regression based loss (i.e. L1 loss) with \textit{soft argmin} operation~\cite{gcnet}, classification based loss with the offset module (i.e. the model is trained with the cross entropy loss to classify the fixed discrete depths and with the L1 loss to regress the offset values) and our proposed adaptive Wasserstein loss with offset module on DTU evaluation set to verify the effectiveness of adaptive Wasserstein loss. We use the same network architecture and parameters for the three losses and trained on DTU training set for 2 epochs. As shown in Tab.~\ref{tab:dif loss}, the adaptive Wasserstein loss improves the accuracy score from 0.423 to 0.364, the completeness score from 0.408 to 0.354, the overall score from 0.416 to 0.359, respectively. As shown in Fig.~\ref{fig:comparision}, the adaptive Wasserstein loss with the offset module (Fig.~\ref{fig:comparision}(f)) predicts more accurate and complete depth values. The regression based loss (Fig.~\ref{fig:comparision}(b)) performs badly for the challenging regions such
as low-textured planes, occluded areas and boundary areas where the pixels are mismatched and the \textit{expectation} values are in multi-modal distribution with wrong depth prediction as shown in Fig.~\ref{fig:comparision}(g). The adaptive Wasserstein loss with the offset module performs better than the classification based loss with offset module as shown in Fig.~\ref{fig:comparision}(d), (f) and (h). The reason is that Wasserstein distance can effectively measure the divergence between the true and the predicted depth distributions that do not have any common supports. 

\subsubsection{Effectiveness of Offset Module}

Next, in order to verify whether the designed offset module can help to achieve sub-pixel accuracy, we train the network with and without the offset module using the adaptive Wasserstein loss and the cross entropy loss. As shown in Tab.~\ref{tab:offset}, the overall score for model trained with adaptive Wasserstein loss is changed from 0.359 to a larger number of 0.377 without the offset values and the overall score for model trained with cross entropy loss is changed from 0.379 to 0.386 without the offset values. We can also see from Fig.~\ref{fig:comparision}(c) and (d), (e) and (f) and (h) that the offset module can help to improve the accuracy of depth prediction and produce smoother depth maps, demonstrating the effectiveness of the proposed offset module which achieves sub-pixel depth accuracy.

\section{Conclusion}

In this paper, we propose adaptive Wasserstein loss for deep depth based MVS, which is able to narrow down the divergence between the true and predicted depth distributions that may not have any common supports. Moreover, we propose a new neural network architecture for learning based MVS that outputs continuous depth values via an offset module. This simple addition of predicted offsets allows us to use the \textit{mode} as the prediction instead of the \textit{expectation} during inference, so as to ensure that the predicted depth has a high estimation probability. Extensive experiments on DTU, Tanks and Temples and BlendedMVS benchmarks show our method is comparable with previous state-of-the-art methods. Thorough analyses demonstrate the benefits of the introduced adaptive Wasserstein loss and offset module for learning based MVS.

{\small
\bibliographystyle{ieee_fullname}
\bibliography{egbib}
}

\end{document}